\documentclass[conference]{IEEEtran}
\IEEEoverridecommandlockouts
\usepackage{subcaption}
\usepackage{cite}
\usepackage{amsmath,amssymb,amsfonts}
\usepackage{booktabs}
\usepackage{algorithm2e}
\usepackage{algpseudocode}
\usepackage{graphicx}
\usepackage{url}
\usepackage{textcomp}
\usepackage{xcolor}

\def\BibTeX{{\rm B\kern-.05em{\sc i\kern-.025em b}\kern-.08em
    T\kern-.1667em\lower.7ex\hbox{E}\kern-.125emX}}

\begin{document}

\title{MRI Patterns of the Hippocampus and Amygdala for Predicting Stages of Alzheimer’s Progression: A Minimal Feature Machine Learning Framework}


\author{
\IEEEauthorblockN{Aswini Kumar Patra}
\IEEEauthorblockA{\textit{Department of Computer Science and Engineering} \\
\textit{NERIST}\\
Itanagar, India \\
akp@nerist.ac.in, https://orcid.org/0000-0002-6278-6534}
\and
\IEEEauthorblockN{Soraisham Elizabeth Devi}
\IEEEauthorblockA{\textit{Department of Computer Science and Engineering} \\
\textit{NERIST}\\
Itanagar, India \\
soraishamelizabeth2017@gmail.com}
\and
\IEEEauthorblockN{Tejashwini Gajurel}
\IEEEauthorblockA{\textit{Neerja Modi School} \\
Jaipur, Rajasthan\\
gajureltejashwini@gmail.com}
}

\maketitle

\begin{abstract}
Alzheimer's disease (AD) progresses through distinct stages, from early mild cognitive impairment (EMCI) to late mild cognitive impairment (LMCI) and eventually to AD. Accurate identification of these stages, especially distinguishing LMCI from EMCI, is crucial for developing pre-dementia treatments but remains challenging due to subtle and overlapping imaging features. This study proposes a minimal-feature machine learning framework that leverages structural MRI data, focusing on the hippocampus and amygdala as regions of interest. The framework addresses the curse of dimensionality through feature selection, utilizes region-specific voxel information, and implements innovative data organization to enhance classification performance by reducing noise. The methodology integrates dimensionality reduction techniques such as PCA and t-SNE with state-of-the-art classifiers, achieving the highest accuracy of 88.46\%. This framework demonstrates the potential for efficient and accurate staging of AD progression while providing valuable insights for clinical applications.
\end{abstract}




\section{Introduction}
Medical imaging assists physicians for better diagnosis, surgical intervention, treatment and follow-up of diseases, and designing better rehabilitation plans \cite{gulo_techniques_2017}. Analysis of medical images plays a crucial role in carrying out these objectives by integrating systems and techniques based on images
	acquired by different imaging modalities. It involves the extraction, selection, and processing of relevant information accurately and
	consistently with minimal execution time.

Alzheimer's disease(AD) transits through different stages, from early mild cognitive impairment(EMCI) to late MCI (LMCI) to AD. It is challenging to identify LMCI from EMCI because of the subtle changes in characteristics, which could not be readily noticeable. The discriminations between different AD stages are crucial for future pre-dementia treatment. Inspired by the above motivation, our primary focus of the study is to sub-group the subjects based on structural MRI scans that exhibit similar characteristics based on region-specific patterns.

MRI studies have shown a link between the volumes of the hippocampus and amygdala and memory performance in individuals with Alzheimer's disease \cite{basso_volumetry_2006}. Another study has noted the significance of temporal lobe structures such as the hippocampus and amygdala in Alzheimer's disease \cite{coupe_lifespan_2019}. Hence, we opted to focus on the hippocampus and amygdala as our regions of interest (ROIs) for this study.

Alzheimer’s disease (AD) could be described into following four stages: healthy control (HC), early mild cognitive impairment (EMCI), late MCI (LMCI) and AD dementia \cite{bi_analysis_2018}. The discriminations between different stages of AD are considerably important issues for future pre-dementia treatment. However, it is still challenging to identify LMCI from EMCI because of the subtle changes in imaging which are not noticeable. In addition, there were relatively few studies to make inferences about the brain dynamic changes in the cognitive progression from EMCI to LMCI to AD. Inspired by the above problems, our primary focus of the study is to predict various stages of the disease. Machine Learning and deep learning techniques widely used for the disease classification as extensively reviewed by Upadhyay et al.\cite{upadhyay_com_2024}.

We propose a novel data-driven framework to sub-group the subjects by an interpretable machine learning approach. It is unique in four aspects: First, it uses minimal feature space, thereby overcoming the curse of dimensionality. Second, The region-specific voxel information and the proposed input data organization make the model perform relatively better as it filters out the noisy features. Thirdly, the relative importance of the features are determined by making the proposed model interpretable. Fourth, the ensemble nature of the framework brings in actionable insights that could be helpful in clinical practice.

The contributions of the work are as follows:
\begin{itemize}
    \item We utilized minimal features by focusing solely on region-specific areas and effectively organizing them.
    \item We proposed a novel machine learning framework incorporating dimension reduction transformation methods, leading to improved prediction accuracy for the stages of Alzheimer's disease, namely EMCI, LMCI, and AD.
\end{itemize}
 
\section{Material and Methods} 
\subsection{Data Description and Pre-Processing}
The neuro-imaging data used in this study were obtained from the ADNI \cite{ADNI_data} repository. We collected 342 T1-weighted structural magnetic resonance imaging(sMRI) images acquired using sagittal orientation and MPRAGE sequences. The dataset includes 104 AD, 103 LMCI, and 105 EMCI cases, with participants aged 65 to 85 years.  

\subsection{Methodology}
\begin{figure*}[!htp]
    \centering
    \includegraphics[width = 0.6\textwidth]{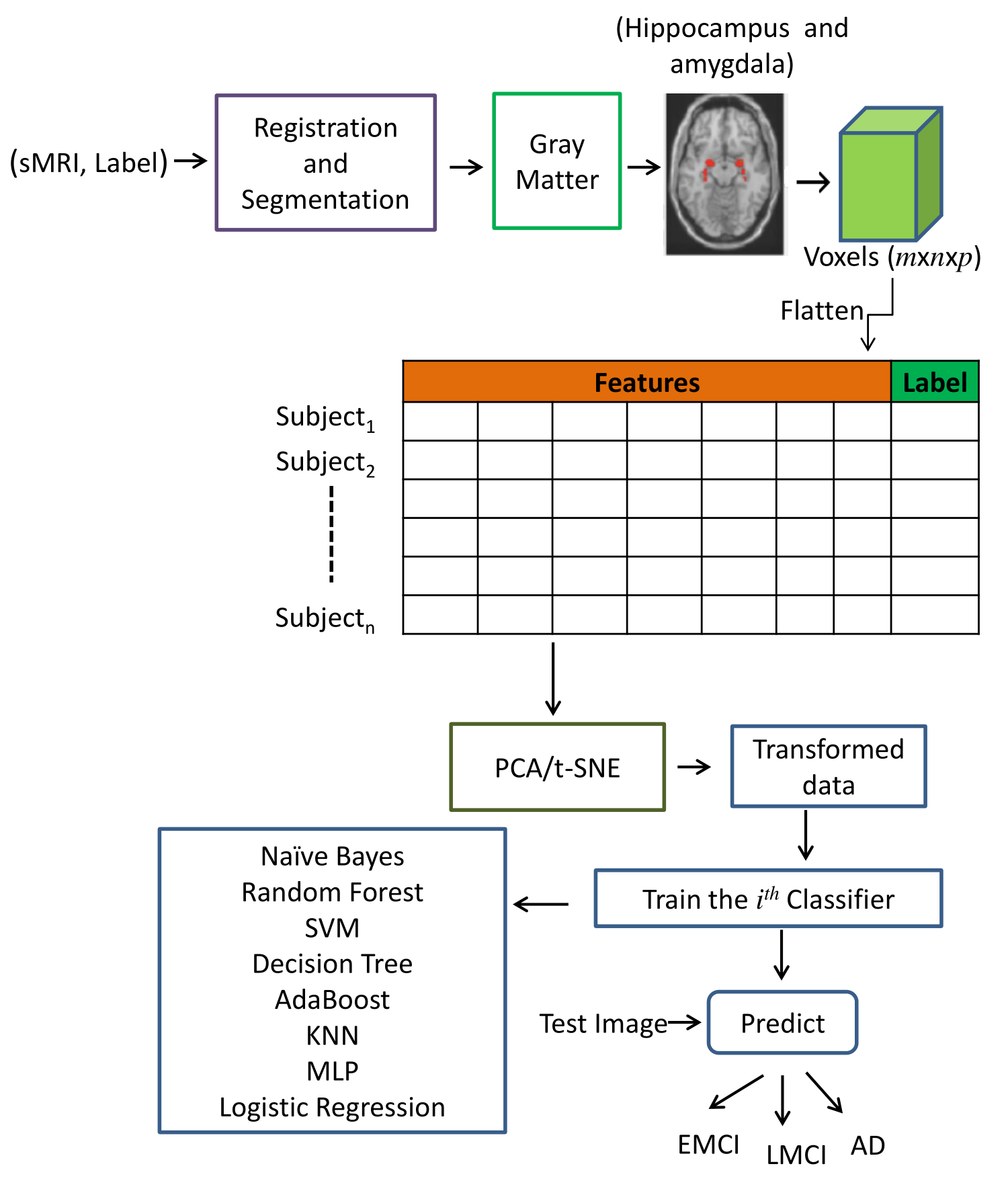}
	\caption{The Proposed Framework}
	\label{fig:block}
\end{figure*}

The framework begins by taking structural MRI (sMRI) scans and their corresponding labels (e.g., EMCI, LMCI, or AD) as input. These MRI images undergo preprocessing steps, including registration to align them in a standard space and segmentation to isolate gray matter, focusing on the hippocampus and amygdala, regions closely associated with cognitive impairment. The gray matter is then represented as a 3D voxel matrix ($m \times n \times p$), flattened into a feature vector for each subject.
Next, the p slices of size $m \times n$ in every subject are flattened to form the feature vector. All feature vector and their corresponding labels are arranged in tabular form. Dimensionality reduction techniques such as Principal Component Analysis (PCA) or t-distributed Stochastic Neighbor Embedding (t-SNE) are applied to transform the high-dimensional feature data into a more compact representation, retaining the most discriminative features.
The transformed data serves as input for the training phase, where various classifiers—including Naïve Bayes, Random Forest, Support Vector Machine (SVM), Decision Tree, AdaBoost, K-Nearest Neighbors (KNN), Multi-Layer Perceptron (MLP), and Logistic Regression—are trained. During testing, a new MRI scan undergoes the same preprocessing and transformation steps, and the trained classifier predicts its label as EMCI, LMCI, or AD. The proposed pipeline outlines a standard machine learning pipeline for classifying sMRI scans into diagnostic categories, involving preprocessing, feature extraction (flattening), dimensionality reduction, training various classifiers, and prediction on new data. The entire framework is illustrated in Fig. \ref{fig:block}.

\subsubsection{Co-registration and Segmentation} Co-registration of the subjects was performed using SPM12 \cite{SPM12}, with a voxel size of 2×2×2 $mm^3$ with  a standard template.

The segmentation process is performed to extract gray tissues using SPM12.
The SPM anatomy toolbox and WFU Pickatlas are used to generate masks of regions of interest. These masks are used to extract the required spatial maps from the segmented images. Then, voxel information are extracted from these spatial maps and organized into a feature matrix, which acts as input to the proposed algorithm.

\subsubsection{Feature Extraction and Organization}

\begin{algorithm*}[!htp]
\caption{Feature Extraction and Organization}
\label{alg:feature}
\KwIn{Atlas mask \( \text{mask} \), List of MRI files \( \text{AD\_subjects} \)}
\KwOut{Feature matrix \( \text{features} \), Corresponding labels \( \text{labels} \)}

\SetKwFunction{FMain}{FeatureExtraction}
\SetKwProg{Fn}{Function}{:}{}
\Fn{\FMain{$\text{mask}, \text{AD\_subjects}$}}{
    \KwData{$\text{mask}$: Brain atlas mask, $\text{AD\_subjects}$: List of MRI file paths}
    
    Initialize $\text{features} \gets []$ \tcp*{Initialize feature matrix}
    Initialize $\text{labels} \gets []$ \tcp*{Initialize labels}
    
    \For{each MRI file \( \text{subject} \) in \( \text{AD\_subjects} \)}{
        \tcp{Step 1: Read MRI data}
        \( \text{filename} \gets \text{subject.name} \)\;
        \( \text{brain} \gets \text{niftiread}(\text{filename}) \)\;
        
        \tcp{Step 2: Apply Mask to Extract Relevant Voxels}
        \( \text{n\_mask} \gets \text{find}(\text{mask}) \)\;
        \( \text{selected\_voxels} \gets \text{brain}(\text{n\_mask}) \)\;
        
        \tcp{Step 3: Reshape Extracted Voxels}
        \( [m, n, p] \gets \text{size}(\text{selected\_voxels}) \)\;
        \( \text{selected\_voxels\_reshaped} \gets \text{reshape}(\text{selected\_voxels}, [m, n, p]) \)\;
        
        \tcp{Step 4: Append Features and Labels}
        \( \text{features} \gets [\text{features}; \text{selected\_voxels\_reshaped}(:)'] \)\;
        \( \text{labels} \gets [\text{labels}; \text{label\_of\_subject}] \)\;
    }
    
    \Return{$\text{features}, \text{labels}$} \tcp*{Return feature matrix and labels}
}
\end{algorithm*}
Structural magnetic resonance imaging studies suggest that specific parts of the brain such as the hippocampus, amygdala, and posterior cingulate, among many others, are affected with the onset of AD\cite{johnson_brain_2012}. It motivated our proposed work to investigate on region-specific features from hippocampus, amygdala for the disease. 
We have considered the SPM anatomy toolbox and WFU Pickatlas \cite{RoI_atlas}  to generate masks of regions of interest. These masks are used to extract the required spatial maps from the segmented images. 

The extracted information from the gray matter spatial map are, in fact, the voxels of our interest. Typically, a dimension $m \times n \times p$ of the  extracted part indicates that it consists of $p$ number of slices of size $p \times q$ each. These 3D information have been arranged as one row vector to represent one subject. In other words, each subject is represented by one row in the feature matrix. The entire feature extraction and organization process is outlined in the algorithm \ref{alg:feature}.

\subsubsection{Dimension Reduction}

Dimensionality reduction can handle large datasets and has multi-fold objectives, namely improving computational efficiency analytical model performance models \cite{jolliffe_principal_2016}. The principal component analysis (PCA) is a popular dimension reduction technique offering several key advantages, particularly for large datasets, by reducing the number of variables while preserving essential information. It transforms a dataset of potentially correlated variables into a smaller set of uncorrelated variables called principal components. They are, in fact, the linear combination of the original variables. These components are ordered by the amount of variance they explain, with the first few capturing most of the variance in the data \cite{abdi_principal_2010}. PCA is also a valuable data visualization and interpretation tool. Lever et al. emphasize its utility in exploratory data analysis, enabling a clearer understanding of the underlying structure within datasets \cite{lever_principal_2017}. The PCA process typically involves several steps: data standardization, scaling the data to have zero mean and unit variance; covariance matrix computation, calculating the covariance matrix of the standardized data; eigenvalue and eigenvector calculation, determining the eigenvalues and eigenvectors of the covariance matrix, where the eigenvectors represent the principal components and the eigenvalues represent the variance explained by each component; and data projection, projecting the original data onto the lower-dimensional space defined by the selected principal components. PCA makes datasets more manageable and interpretable by focusing on the most informative features and minimizing redundancy.

t-Distributed Stochastic Neighbor Embedding (t-SNE) is a non-linear dimensionality reduction technique designed primarily for visualizing high-dimensional datasets in a low-dimensional space, such as two or three dimensions. t-SNE works by preserving the local structure of data, focusing on capturing pairwise similarities between points in the high-dimensional and low-dimensional spaces \cite{maaten_visualizing_2008}.
Unlike linear techniques like PCA, t-SNE is particularly effective at uncovering complex relationships in data with non-linear structures. It minimizes the divergence between probability distributions representing these pairwise similarities in the original and reduced spaces. The method places similar data points closer in the visualization, forming clusters that often correspond to meaningful patterns or groupings \cite{maaten_visualizing_2008}.
However, due to its computational complexity, it is primarily a visualization tool and should not be used for tasks requiring predictive modeling or scalability to huge datasets \cite{wattenberg2016how}. Despite its limitations, t-SNE remains a powerful tool for identifying patterns and relationships in complex, high-dimensional data.

\subsubsection{Classification}
 We considered eight widely used machine learning classifiers to benchmark the proposed approach, each bringing complementary strengths to the evaluation process. Support Vector Machines (SVM) are robust classifiers that maximize the margin between classes, making them particularly effective in high-dimensional spaces. Decision Trees are versatile models that split data based on feature thresholds, creating a tree-like structure for decision-making, and are well-suited for capturing non-linear relationships despite a tendency to overfit. Multi-layer perceptrons (MLP), as neural networks, utilize multiple layers of interconnected nodes to capture complex, non-linear patterns in data. Grounded in Bayes' theorem, Naive Bayes classifiers assume conditional independence between features and excel in tasks like text classification. K-Nearest Neighbors (KNN), a simple instance-based learning method, classifies data based on the majority label among its nearest neighbors. AdaBoost, an ensemble method, iteratively combines weak learners to improve accuracy by focusing on difficult-to-classify samples. Random Forests enhance robustness and mitigate over-fitting by aggregating predictions from multiple decision trees. Finally, Logistic Regression is a simple yet effective probabilistic model that predicts class membership by estimating the likelihood of a data point belonging to a class based on linear combinations of features.

These transformed feature sets are then integrated with various classifiers to effectively distinguish between different stages of the disease. The complete workflow for this process is systematically described in Algorithm \ref{alg:classify}.

\begin{algorithm*}[!htp]
\caption{Dimensionality Reduction and Classification}
\label{alg:classify}
\KwIn{Feature matrix \( X \), Corresponding target labels \( y \), Number of principal components \( n \), List of classifiers \( \{C_1, C_2, \ldots, C_m\} \)}
\KwOut{Best combination of dimensionality reduction technique and classifier, along with mean accuracy}

\SetKwFunction{FMain}{DimensionalityReductionClassification}
\SetKwProg{Fn}{Class}{:}{}
\Fn{\FMain{$X, y, n, \{C_1, C_2, \ldots, C_m\}$}}{
    \KwData{$X$: Feature matrix, $y$: Target labels, $n$: Number of principal components, \( \{C_1, C_2, \ldots, C_m\} \): List of classifiers}
    
    \tcp{Step 1: Apply PCA Transformation}
    Initialize PCA with \( n \) components\;
    \( X_{\text{pca}} \gets \text{PCA}(n_{\text{components}}=n).fit\_transform(X) \)\;
    
    \tcp{Step 2: Apply t-SNE Transformation}
    Initialize t-SNE with \( n \) components\;
    \( X_{\text{tsne}} \gets \text{TSNE}(n_{\text{components}}=n).fit\_transform(X) \)\;
    
    \tcp{Step 3: Evaluate PCA + Classifiers}
    \For{each classifier \( C \) in \( \{C_1, C_2, \ldots, C_m\} \)}{
        \tcp{Initialize list to store accuracies for each fold}
        Initialize \( \text{accuracies} \gets [] \)\;
        
        \For{each fold in \( \text{kf.split}(X_{\text{pca}}, y) \)}{
            \tcp{Train and test the classifier}
            Train \( C \) on training data\;
            Test \( C \) on validation data\;
            Record accuracy\;
        }
        
        \tcp{Store the mean accuracy of the classifier}
        \( \text{mean\_accuracy\_pca}[C] \gets \text{mean}(\text{accuracies}) \)\;
    }
    
    \tcp{Step 4: Evaluate t-SNE + Classifiers}
    \For{each classifier \( C \) in \( \{C_1, C_2, \ldots, C_m\} \)}{
        \tcp{Initialize list to store accuracies for each fold}
        Initialize \( \text{accuracies} \gets [] \)\;
        
        \For{each fold in \( \text{kf.split}(X_{\text{tsne}}, y) \)}{
            \tcp{Train and test the classifier}
            Train \( C \) on training data\;
            Test \( C \) on validation data\;
            Record accuracy\;
        }
        
        \tcp{Store the mean accuracy of the classifier}
        \( \text{mean\_accuracy\_tsne}[C] \gets \text{mean}(\text{accuracies}) \)\;
    }
    
    \tcp{Step 5: Select Best Combination}
    \( \text{best\_combination} \gets \text{argmax}(\text{mean\_accuracy\_pca} \cup \text{mean\_accuracy\_tsne}) \)\;
    
    \Return{Best combination of dimensionality reduction technique and classifier, with corresponding mean accuracy}\;
}
\end{algorithm*}

\section{Results and Discussion}
We have investigated the classification of EMCI, LMCI, and AD stages through three steps: 1) principal component analysis (PCA) and t-SNE analysis of the proposed minimal feature set derived from the hippocampus and amygdala, 2) machine learning classifiers utilizing the original features PCA transformation, and t-SNE transformation and 3) Choosing the best combination resulting in improved classification accuracy.

We utilized the \textit{sklearn} package in Python for performing principal component analysis (PCA) and t-SNE for dimensionality reduction, as well as for implementing a range of classifiers and evaluating their performance using various metrics. The experiments were conducted on the Kaggle platform.

\subsection{PCA and t-SNE based Transformation} 
\begin{figure}[hbt!]
    \centering
    \begin{subfigure}{0.45\textwidth}
        \centering
        \includegraphics[width=1\textwidth]{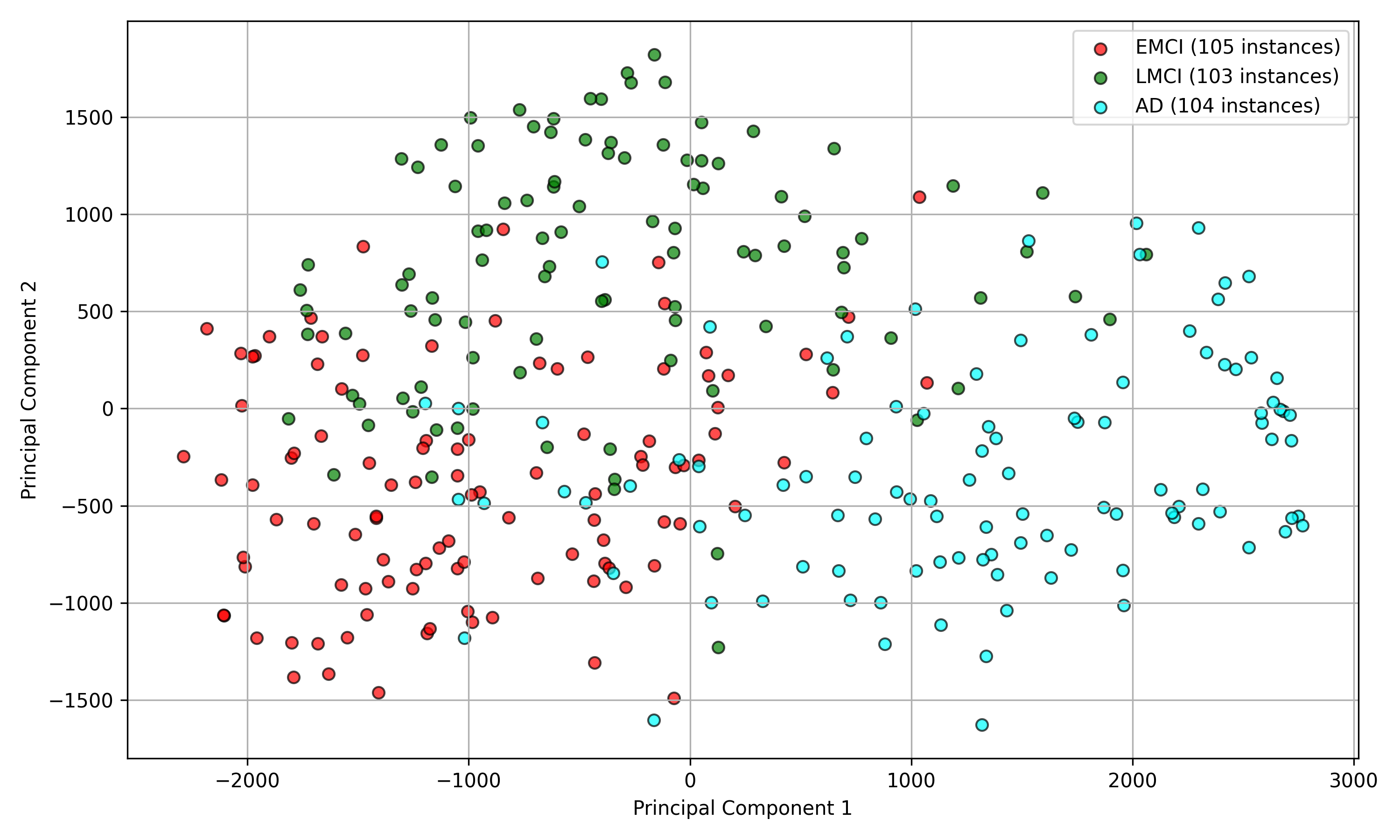}
        \caption{}
        \label{fig:pca_original}
    \end{subfigure}
    \begin{subfigure}{0.45\textwidth}
        \centering
        \includegraphics[width=1\textwidth]{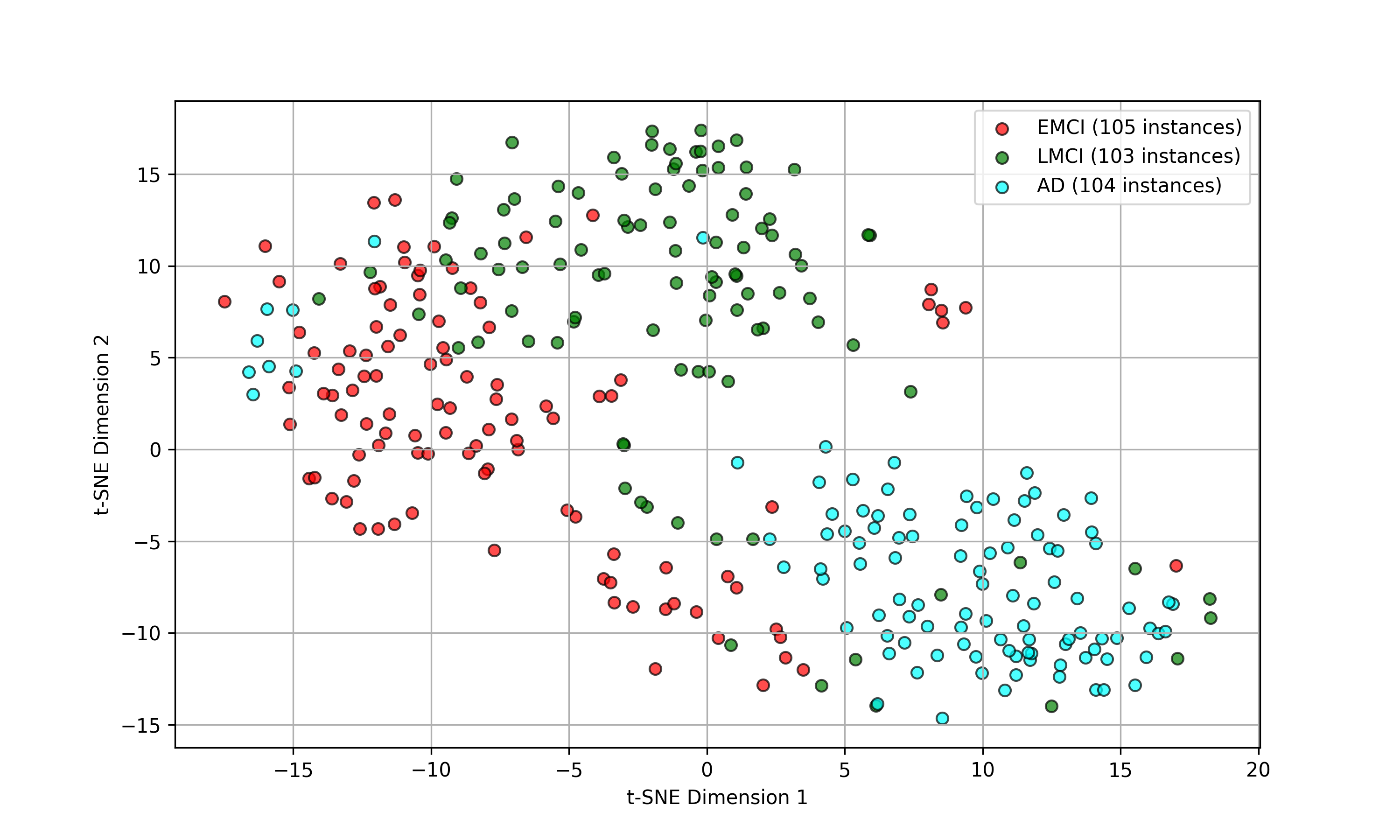}
        \caption{}
        \label{fig:tsne_org}
    \end{subfigure}
    \caption{\subref{fig:pca_original}) PCA projection of classes  and \subref{fig:tsne_org}) t-SNE Embeddings}
    \label{fig:pca}
\end{figure}

The visualization in Fig. \ref{fig:pca_original} highlights the data projected onto two principal components using Principal Component Analysis (PCA). It displays three categories of data: EMCI, LMCI, and AD (Alzheimer's Disease). Each category forms distinguishable clusters, although there is a noticeable overlap among them. The EMCI data points, represented in red, are primarily concentrated on the negative side of Principal Component 1, whereas the AD points, in cyan, are more dispersed towards the positive side. The LMCI points, in green, are distributed more centrally, overlapping with both the EMCI and AD categories. This visualization suggests that the features used in PCA capture meaningful differences between the groups, although the overlap hints at potential challenges in achieving perfect separability.
We also used a two-dimensional projection of data with t-SNE (t-distributed Stochastic Neighbor Embedding), which presents the EMCI data points (red) and AD data points (cyan) forming more distinct clusters with fewer overlaps compared to the PCA plot. However, the LMCI data points (green) remain centrally distributed, overlapping with the other two categories, as shown in Fig. \ref{fig:tsne_org}. This analysis could be further complemented by applying classification techniques to evaluate and enhance the distinction among the groups.

\subsection{Comparison of Various Combinations}
Table \ref{tab:classifier_performance} presents a comparison of different machine learning classifiers-SVM, Decision Tree (DT), Multi-Layer Perceptron (MLP), Naive Bayes (NB), K-Nearest Neighbors (KNN), AdaBoost, Random Forest (RF), and Logistic Regression (LR)-evaluating their performance with and without data transformation techniques such as PCA (Principal Component Analysis) and t-SNE (t-Distributed Stochastic Neighbor Embedding). The performance metrics, including precision, recall, F1-score, and accuracy, are reported as weighted values due to the use of 5-fold cross-validation, ensuring that the results account for class imbalances and provide a more robust performance measure.
The results indicate that PCA generally improves the performance of classifiers such as SVM, KNN, and Random Forest, showing higher precision, recall, F1-score, and accuracy compared to their base models. Notably, KNN benefits significantly from PCA, exhibiting strong performance across all metrics. On the other hand, t-SNE, while useful for dimensionality reduction, does not offer substantial improvements and often results in lower performance compared to PCA for most classifiers. For instance, the precision and recall values for SVM and KNN are higher when combined with PCA, but the performance with t-SNE tends to be less favorable. Overall, PCA proves to be a more effective transformation technique for enhancing the performance of these classifiers, while t-SNE seems less beneficial in this context.

\begin{table*}[!htbp]
\centering
\caption{Performance Comparison of Different Classifiers}
\label{tab:classifier_performance}
\begin{tabular}{lcccccc}
\toprule
\textbf{Classifier}           & \textbf{Transformation}       & \multicolumn{1}{c}{\textbf{Precision}} & \multicolumn{1}{c}{\textbf{Recall}} & \multicolumn{1}{c}{\textbf{F1-Score}} & \textbf{Accuracy} \\
                              &                               & \multicolumn{1}{c}{\textbf{(Weighted)}} & \multicolumn{1}{c}{\textbf{(Weighted)}} & \multicolumn{1}{c}{\textbf{(Weighted)}} &                     \\
\midrule
SVM                           & SVM                      & 0.8729                                & 0.8687                                & 0.8691                                  & 0.8687              \\
                              & PCA+SVM                       & 0.8776                                & 0.8718                                & 0.8726                                  & 0.8718              \\
                              & t-SNE+SVM                     & 0.8321                                & 0.8303                                & 0.8281                                  & 0.8303              \\
\midrule
Decision Tree                 & DT                      & 0.7240                                & 0.7022                                & 0.7055                                  & 0.7022              \\
                              & PCA+DT                        & 0.7442                                & 0.7402                                & 0.7397                                  & 0.7402              \\
                              & t-SNE+DT                      & 0.7844                                & 0.7788                                & 0.7788                                  & 0.7788              \\
\midrule
MLP                           & MLP                      & 0.8695                                & 0.8656                                & 0.8656                                  & 0.8656              \\
                              & PCA+MLP                       & 0.8433                                & 0.8399                                & 0.8400                                  & 0.8399              \\
                              & t-SNE+MLP                     & 0.8010                                & 0.7981                                & 0.7968                                  & 0.7981              \\
\midrule
Naive Bayes                   & NB                      & 0.8024                                & 0.7884                                & 0.7886                                  & 0.7884              \\
                              & PCA+NB                        & 0.8299                                & 0.8207                                & 0.8218                                  & 0.8207              \\
                              & t-SNE+NB                      & 0.7651                                & 0.7596                                & 0.7567                                  & 0.7596              \\
\midrule
KNN                           & KNN                      & 0.8419                                & 0.8268                                & 0.8253                                  & 0.8268              \\
                              & PCA+KNN                       & 0.8887                                & 0.8846                                & 0.8847                                  & 0.8846              \\
                              & t-SNE+KNN                     & 0.8569                                & 0.8526                                & 0.8518                                  & 0.8526              \\
\midrule
AdaBoost                      & AdaBoost                     & 0.8389                                & 0.8175                                & 0.8201                                  & 0.8175              \\
                              & PCA+AdaBoost                  & 0.7360                                & 0.7054                                & 0.7055                                  & 0.7054              \\
                              & t-SNE+AdaBoost                & 0.7859                                & 0.7664                                & 0.7679                                  & 0.7664              \\
\midrule
Random Forest                 & Random Forest                      & 0.8511                                & 0.8432                                & 0.8444                                  & 0.8432              \\
                              & PCA+RF                        & 0.8409                                & 0.8336                                & 0.8351                                  & 0.8336              \\
                              & t-SNE+RF                      & 0.8319                                & 0.8269                                & 0.8255                                  & 0.8269              \\
\midrule
Logistic Regression           & LR                      & 0.8606                                & 0.8589                                & 0.8589                                  & 0.8589              \\
                              & PCA+LR                        & 0.8071                                & 0.8012                                & 0.8021                                  & 0.8012              \\
                              & t-SNE+LR                      & 0.7784                                & 0.7758                                & 0.7748                                  & 0.7758              \\
\bottomrule
\end{tabular}
\end{table*}

\subsection{Class-wise Performance}
\begin{figure*}[!htbp]
    \centering
    \begin{subfigure}{0.45\textwidth} 
        \centering
        \includegraphics[scale=0.45]{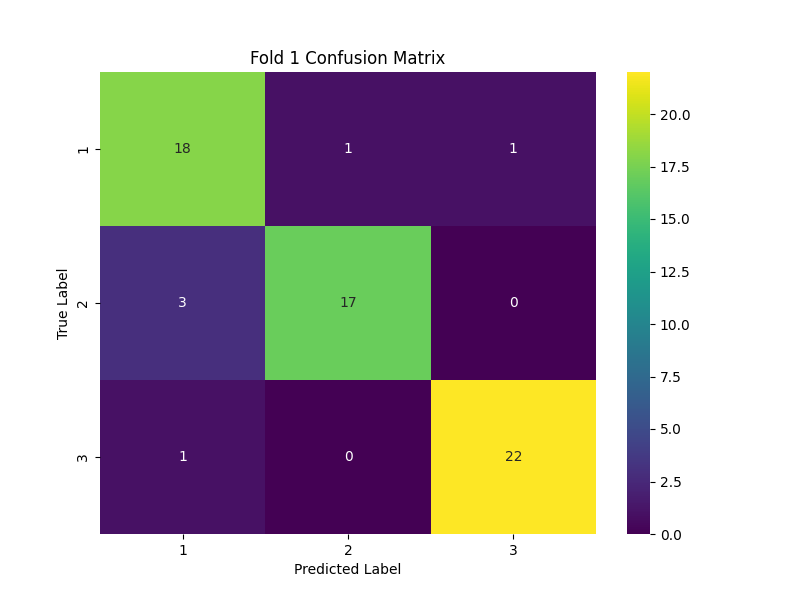}
        \caption{}
        \label{fig:confusion_matrix_fold_1}
    \end{subfigure}
    \hfill
    \begin{subfigure}{0.45\textwidth} 
        \centering
        \includegraphics[scale=0.45]{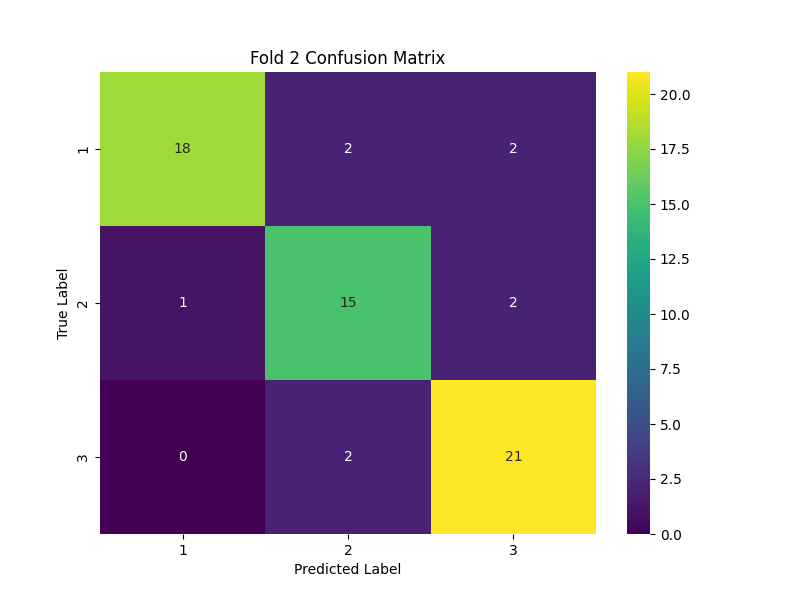}
        \caption{}
        \label{fig:confusion_matrix_fold_2}
    \end{subfigure}

    \vspace{0.5em} 

    \begin{subfigure}{0.45\textwidth}
        \centering
        \includegraphics[scale=0.45]{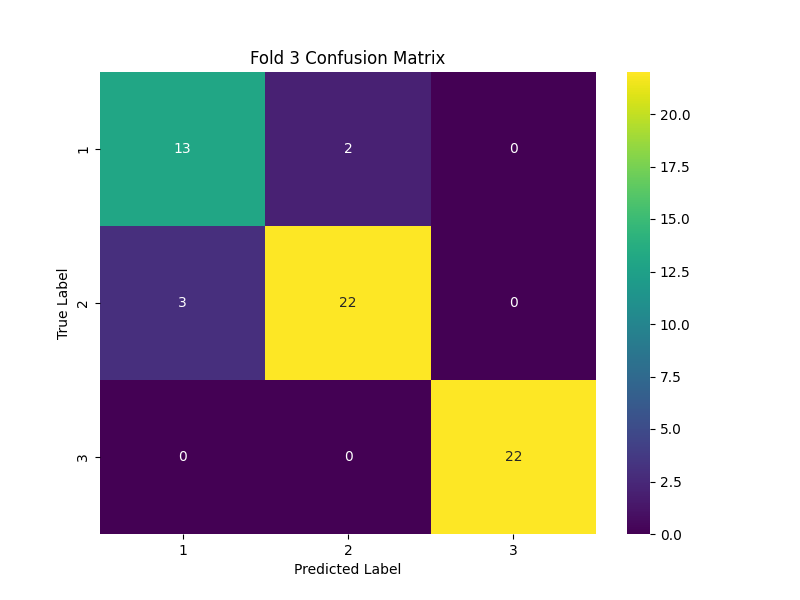}
        \caption{}
        \label{fig:confusion_matrix_fold_3}
    \end{subfigure}
    \hfill
    \begin{subfigure}{0.45\textwidth}
        \centering
        \includegraphics[scale=0.45]{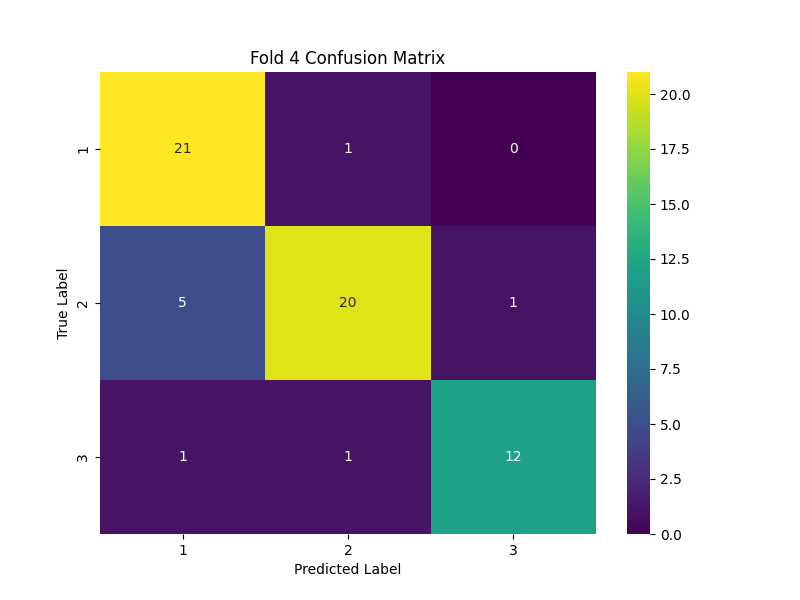}
        \caption{}
        \label{fig:confusion_matrix_fold_4}
    \end{subfigure}

    \vspace{0.5em} 

    \begin{subfigure}{0.45\textwidth}
        \centering
        \includegraphics[scale=0.45]{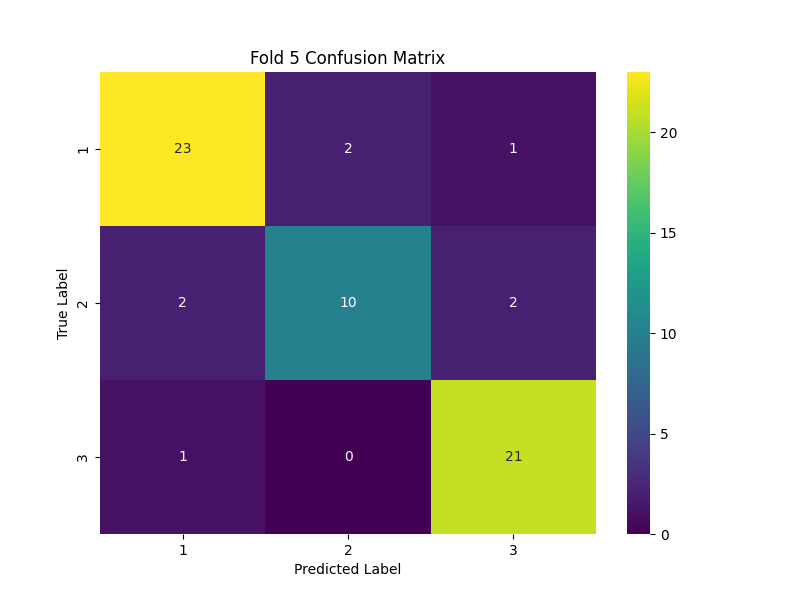}
        \caption{}
        \label{fig:confusion_matrix_fold_5}
    \end{subfigure}

    \caption{Confusion Matrices for Five-Fold Cross-Validation}
    \label{fig:confusion_matrices_folds}
\end{figure*}

\begin{figure}[!hbp]
    \centering
    \begin{subfigure}{0.3\textwidth}
        \centering
        \includegraphics[width=\textwidth]{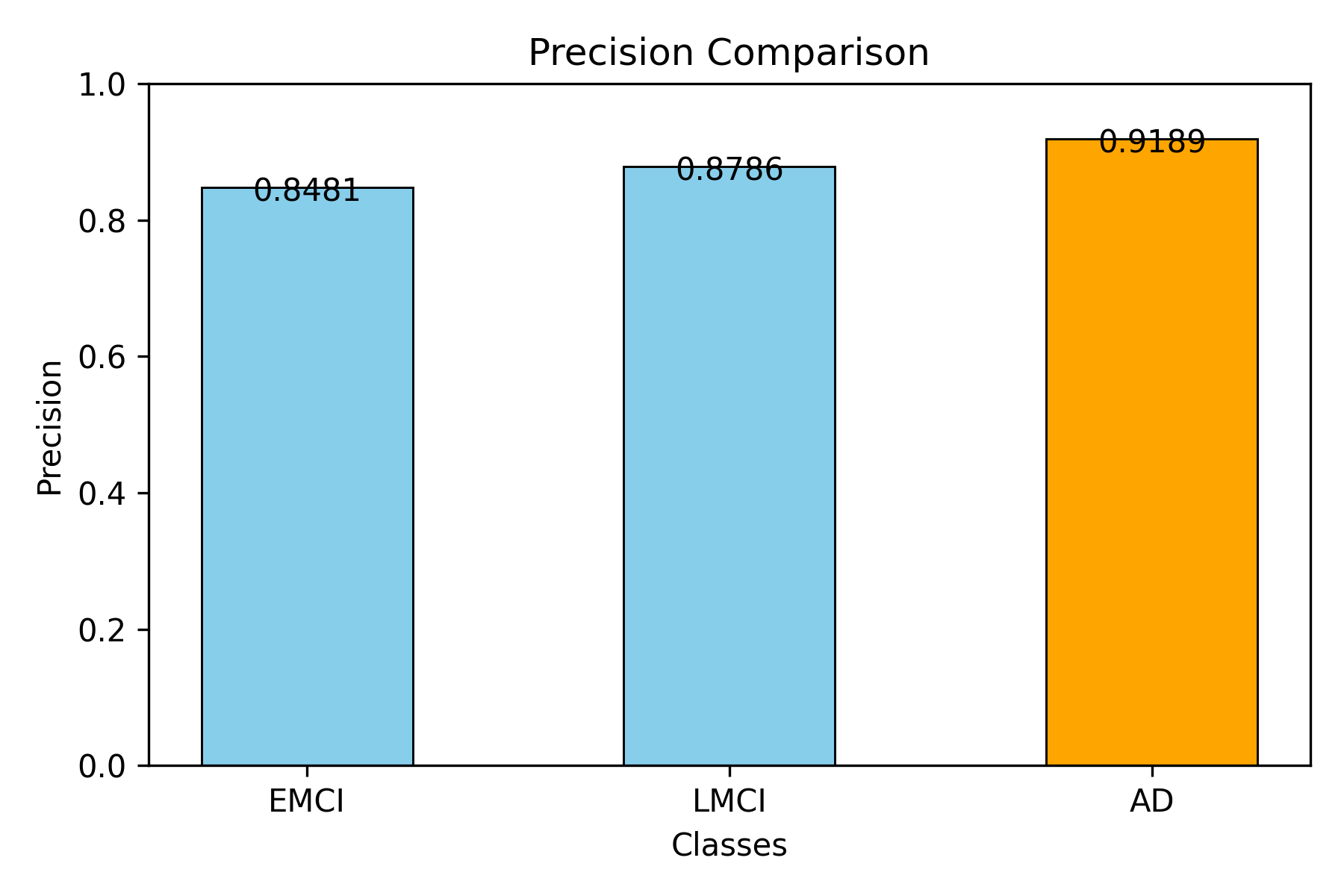} 
        \caption{Precision Comparison}
        \label{fig:precision_comparison}
    \end{subfigure}
    \hfill
    \begin{subfigure}{0.3\textwidth}
        \centering
        \includegraphics[width=\textwidth]{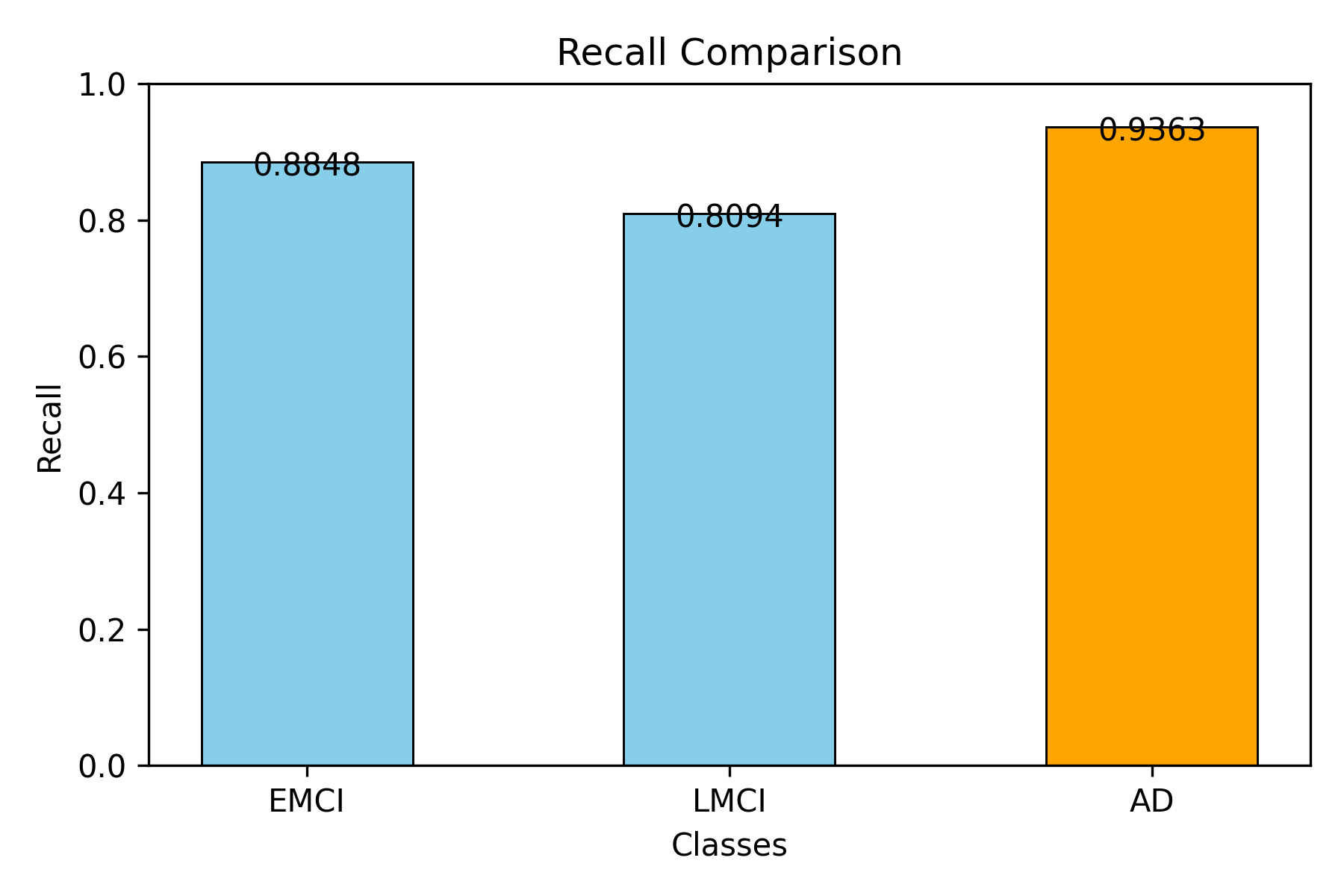} 
        \caption{Recall Comparison}
        \label{fig:recall_comparison}
    \end{subfigure}
    \hfill
    \begin{subfigure}{0.3\textwidth}
        \centering
        \includegraphics[width=\textwidth]{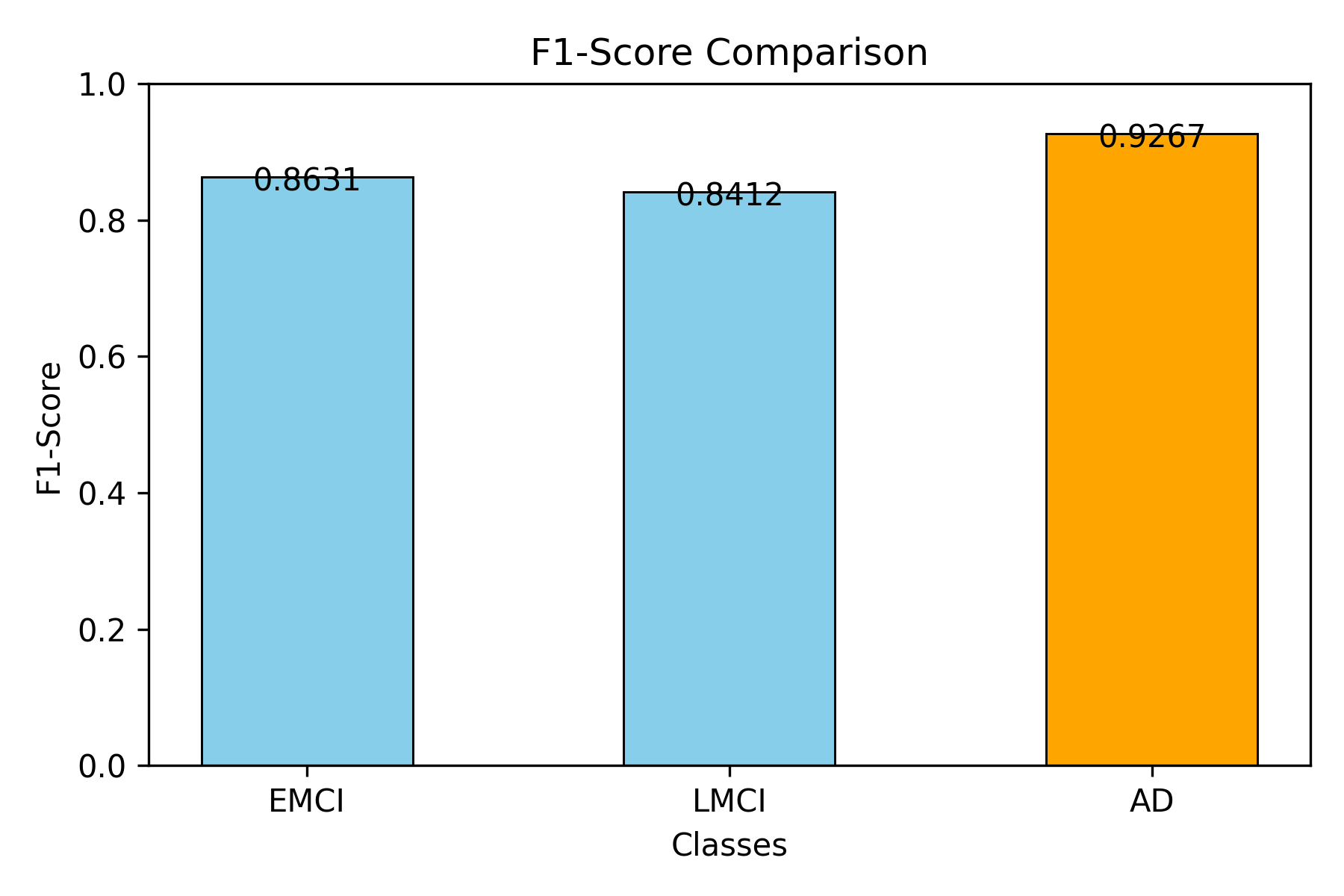} 
        \caption{F1-Score Comparison}
        \label{fig:f1_score_comparison}
    \end{subfigure}

    \caption{Class-wise Performance: Precision, Recall, and F1-Score}
    \label{fig:class_performance}
\end{figure}

PCA + KNN performs the best among the tested configurations, with the evaluation based on five-fold cross-validation results. The confusion matrices for each class (EMCI, LMCI, AD) are shown in Fig. \ref{fig:confusion_matrices_folds}.
For EMCI, the model balances precision (0.8481) and recall (0.8848), resulting in a solid F1-score of 0.8631. Although the model identifies most EMCI cases accurately (indicated by high recall), the slightly lower precision suggests some false positives, i.e., non-EMCI cases being incorrectly classified as EMCI. This implies that while the model effectively detects EMCI, further tuning could help reduce misclassifications.
In the case of LMCI, the model shows a relatively high precision (0.8786), but the recall drops to 0.8094. This indicates that the model misses about 19\% of actual LMCI cases (false negatives), even though it does not frequently misclassify non-LMCI cases as LMCI. The F1-score of 0.8412 reflects this imbalance, suggesting that the model could benefit from optimization, particularly in improving recall for LMCI.
The model achieves the best performance for AD, with high precision (0.9189) and recall (0.9363), resulting in an impressive F1-score of 0.9267. This suggests that the model is highly effective at both detecting AD cases and minimizing false positives. The high recall indicates that most AD cases are identified, while the high precision ensures accurate predictions. Consequently, AD classification is the most reliable among the three. The class-wise comparison of all the metrics are depicted in Fig. \ref{fig:class_performance}
In summary, the model performs optimally for AD, but there is room for improvement in EMCI, particularly in reducing false positives. LMCI, however, exhibits a notable weakness in recall, where many actual LMCI cases are missed. Addressing this issue through further model tuning or feature enhancement could improve the overall robustness of the classifier and make it more reliable for real-world applications.

\section{Conclusion}
This work presents a minimal feature machine learning framework that utilizes region-specific structural MRI patterns. Its outstanding performance stems from integrating dimensionality reduction techniques such as PCA or t-SNE with classifiers, allowing for the effective learning and identification of patterns in the hippocampus and amygdala. Additionally, the model successfully utilizes fewer MRI scans focusing on gray matter, enhancing its efficiency. The study's findings suggest that the combination of PCA and KNN holds promise as a practical approach for accurately classifying patients at various stages of disease progression.
Future research could explore other tissue patterns and increase the number of subjects to validate the model's effectiveness further. This approach has potential applications in clinical practice and diagnosis, offering a streamlined and accurate method for disease assessment.

	\bibliographystyle{unsrt}
	\bibliography{ref_aps}
\end{document}